\documentclass{article}
\usepackage{spconf,graphicx}
\usepackage{amsmath,amsfonts,amssymb,bm}
\usepackage{tabularx,setspace}
\usepackage[ruled]{algorithm2e}
\usepackage{cite}
\usepackage{microtype}
\usepackage{hyperref,color}

\newcommand{\argmin}{\operatornamewithlimits{argmin}}

\title{Heterogeneous Self-Supervised Acoustic Pre-Training with Local Constraints}

\name{Xiaodong Cui$^{1}$, A F M Saif$^{2}$, Brian Kingsbury$^{1}$, Tianyi Chen$^{3}$}
\address{
$^{1}$IBM Research, IBM T. J. Watson Research Center, Yorktown Heights, New York, USA \\
$^{2}$Rensselaer Polytechnic Institute, Troy, New York, USA \\
$^{3}$Cornell Tech, New York, New York, USA
}

\begin{document}
\ninept
\maketitle
\begin{abstract}
Self-supervised pre-training using unlabeled data is widely used in automatic speech recognition. In this paper, we propose a new self-supervised pre-training approach to dealing with heterogeneous data. Instead of mixing all the data and minimizing the averaged global loss in the conventional way, we impose additional local constraints to ensure that the model optimizes each source of heterogeneous data to its local optimum after $K$-step gradient descent initialized from the model. We formulate this as a bilevel optimization problem, and use the first-order approximation method to solve the problem. We discuss its connection to model-agnostic meta learning. Experiments are carried out on self-supervised pre-training using multi-domain and multilingual datasets, demonstrating that the proposed approach can significantly improve the adaptivity of the self-supervised pre-trained model for the downstream supervised fine-tuning tasks.
\end{abstract}
\vspace{-0.2cm}
\section{Introduction}
\label{sec:intro}

Since labeled data is expensive to obtain while unlabeled data is readily available, a common practice in machine learning is a two-stage approach where a large amount of unlabeled data is first used for self-supervised pre-training and the pre-trained foundation model is then fine-tuned using labeled data in downstream tasks. In speech related applications, self-supervised pre-training has been actively investigated and broadly used \cite{baevski2019vq,baevski2020wav2vec,hsu2021hubert,Chiu_Bestrq}.

When carrying out self-supervised acoustic pre-training using unlabeled data, it is inevitable to deal with data heterogeneity as the large amount of training data may come from various sources (e.g., domains and languages) bearing different acoustic characteristics. The conventional approach to self-supervised pre-training mixes all data together and minimizes the averaged loss. A drawback of this strategy is that a low average global loss does not guarantee a low loss for each source of the heterogeneous data. In this paper, we propose a new self-supervised pre-training approach. In this approach, in addition to the averaged global loss across all heterogeneous data sources, we also impose constraints that require each source of the heterogeneous data to reach its local loss optimum after $K$-step gradient descent initialized from the model. Such constraints on the local optimum will ensure that, when optimizing the averaged loss, a good feature representation for each source is also preserved. We formulate this self-supervised pre-training with local constraints (PTLOC) as a bilevel optimization problem \cite{liu2021investigating,crockett2022bilevel,chen2023learning,lu2023meta,franceschi2018bilevel} where the upper problem is the averaged global loss across all data sources, while the lower problem is the local loss of each data source. We use the first-order approximation method to solve this bilevel optimization problem and discuss its connection with model-agnostic meta-learning (MAML) \cite{finn2017model}. We build pre-trained acoustic models using the proposed PTLOC approach and conduct downstream automatic speech recognition (ASR) tasks on two scenarios. One scenario uses speech data from multiple domains and the other uses multilingual speech data. We compare the performance of PTLOC with that of the conventional self-supervised pre-trained models. Our experiments show that the proposed PTLOC can give rise to a better pre-trained model with superior adaptivity in downstream ASR fine-tuning (FT) tasks.

The remainder of the paper is organized as follows.  We formulate the problem of PTLOC in Section \ref{sec:form}.  Its optimization in given in Section \ref{sec:opt} and the pseudo-code implementation is given in Section \ref{sec:impl}. The experimental results on multi-domain and multilingual ASR are reported in Section \ref{sec:exp}. Finally, we conclude the paper with a summary in Section \ref{sec:sum}.

\vspace{-0.2cm}

\section{Problem Formulation}
\label{sec:form}

Suppose the unlabeled data $\mathcal{D}$ is collected from $M$ sources: $\mathcal{D} = \{ \mathcal{D}_{1},\cdots, \mathcal{D}_{M}\}$. Conventional self-supervised learning (CSSL) trains a model that minimizes the following empirical risk
\vspace{-0.2cm}
\begin{align}
   \min_{\theta} \frac{1}{M}\sum_{i=1}^{M}\sum_{x \in \mathcal{D}_{i}}\ell_{i}(\theta; x, \mathcal{D}_{i}) \label{eqn:risk}
\end{align}
where $x\in\mathcal{D}_{i}$ is a sample from source $\mathcal{D}_{i}$; $\theta$ is the model parameters; $\ell_{i}(\theta; x, \mathcal{D}_{i})$ is the loss defined on data source $\mathcal{D}_{i}$. In CSSL, the objective function in Eq. \eqref{eqn:risk} simply mixes heterogeneous data from various sources together. Since each source of data may bear its unique characteristics in feature representations, we want the model to preserve these characteristics to offer more robustness and generalization after pre-training. However, this is not guaranteed in Eq. \eqref{eqn:risk} by only optimizing towards the averaged global loss. Also considering that the pre-trained model will serve as an initialization point for downstream tasks, we are interested in its adaptivity after multiple steps of gradient descent. To that end, we propose PTLOC, where local constraints for each data source are imposed in addition to the global averaged loss, requiring that the model also optimize each source of heterogeneous data to its local optimum after $K$-step gradient descent initialized from $\theta$. This is formulated as the following bilevel optimization problem:
\begin{align}
 & \min_{\theta}\frac{1}{M}\sum_{i=1}^{M} \sum_{x \in \mathcal{D}_{i}}\ell_{i}(\phi^{*}_{i}(\theta);x,\mathcal{D}_{i})  \nonumber \\
 &{\rm s. t.}  \ \ \ \ \phi^{*}_{i}(\theta) = \argmin_{\phi_{i}}\sum_{x \in \mathcal{D}_{i}}\ell_{i}(\phi_{i}(\theta);x,\mathcal{D}_{i}), \ \ i \in [M].   \label{eqn:risk_maml}
\end{align}
where the upper level problem is the averaged global loss with $\theta$ being the initialization model parameter shared by data from all sources and the lower level problem is $M$ local losses with model parameter $\phi_{i}(\theta)$ for each data source. The dependency of $\phi_{i}(\theta)$ on $\theta$ is through a function of $K$-step gradient descent starting from $\theta$, which will become clear in Section \ref{sec:opt}. By imposing the constraints in the lower-level problem, we ensure that the resulting model not only produces a good average global loss but also serves as a good initialization that can reach a local optimal point of the local loss from each data source after a few steps of gradient descent.

\vspace{-0.2cm}
\section{Optimization}
\label{sec:opt}

To solve the bilevel optimization problem in Eq. \eqref{eqn:risk_maml}, we first define the following functions to simplify the derivation
\vspace{-0.2cm}
\begin{align}
     f(\theta)  & \triangleq  \frac{1}{M}\sum_{i=1}^{M} \sum_{x \in \mathcal{D}_{i}}\ell_{i}(\phi^{*}_{i}(\theta);x,\mathcal{D}_{i})  \\
     g_{i}(\phi_{i}) & \triangleq  \sum_{x \in \mathcal{D}_{i}}\ell_{i}(\phi_{i}(\theta);x,\mathcal{D}_{i}), \ \ i \in [M]
\end{align}
where $f(\cdot)$ and $g_{i}(\cdot)$ denote the upper level and lower level problems in Eq. \ref{eqn:risk_maml} respectively.

\vspace{-0.2cm}

\subsection{Solving the lower level problem}
\label{sec:lowerp}

There are $M$ lower-level problems, one for each data source. We first solve each of the lower-level problems $g_{i}(\phi_{i})$ using K-step gradient descent starting from a common parameter $\theta$ shared by all data sources from the upper level, with a learning rate $\alpha$
\begin{align}
    \phi^{i}_{k} = \phi^{i}_{k-1} - \alpha \nabla_{\phi_{i}} g_{i}(\phi^{i}_{k-1}), \ \ i \in [M] \label{eqn:maml_sgd}
\end{align}
where $\phi^{i}_{0} = \theta$. $\phi^{i}_{K}$ is used to approximate $\phi^{*}_{i}(\theta)$: $\phi^{*}_{i}(\theta) \approx \phi^{i}_{K}$.
Particularly, when $K=1$, we have
\begin{align}
      \phi^{*}_{i}(\theta) \approx  \theta - \alpha \nabla_{\phi_{i}} g_{i}(\theta).  \label{eqn:maml_sgd_k1}
\end{align}
The dependency of $\phi_{i}(\theta)$ on $\theta$ is obvious in Eq. \ref{eqn:maml_sgd} with Eq. \eqref{eqn:maml_sgd_k1} being a special case when $K=1$.

\vspace{-0.2cm}

\subsection{Solving the upper level problem}
\label{sec:upperp}

The upper level problem $f(\theta)$ is the global loss of all data sources starting from a shared parameter $\theta$. We also solve it using gradient descent with a learning rate $\beta$
\vspace{-0.2cm}
\begin{align}
   \theta_{t} = \theta_{t-1} - \beta \nabla_{\theta} f(\theta_{t-1}).
\end{align}
Its gradient $\nabla_{\theta} f(\theta)$ is computed based on gradient unrolling \cite{Shaban_BLGradUnrolling,Liu_BLGradUnrolling,Shen_BLGradUnrolling} as follows \footnote{To avoid cluttered notation, we will drop the step index $t$ in $\theta_{t}$ and $\theta$ in $\phi_{i}(\theta)$ and $\phi^{*}_{i}(\theta)$ in the derivation.}
\begin{align}
\nabla_{\theta} f(\theta) &  = \frac{1}{M}\sum_{i=1}^{M} \nabla_{\theta} \left( \sum_{x \in \mathcal{D}_{i}}\ell_{i}(\phi^{*}_{i}(\theta);x,\mathcal{D}_{i}) \right) \nonumber \\
                          &  = \frac{1}{M}\sum_{i=1}^{M} \nabla_{\theta} g_{i}(\phi^{*}_{i}(\theta)) \approx \frac{1}{M}\sum_{i=1}^{M} \nabla_{\theta} g_{i}(\phi^{i}_{K})  \label{eqn:ulthetagrad}
\end{align}
The last step is because $\phi^{*}_{i}(\theta)$ is approximated by  $\phi^{i}_{K}$.

Applying the chain rule, we have
\begin{align}
\nabla_{\theta} f(\theta) &  \approx \frac{1}{M}\sum_{i=1}^{M} \nabla_{\theta} g_{i}(\phi^{i}_{K})  = \frac{1}{M}\sum_{i=1}^{M} \nabla_{\phi_{i}} g_{i}(\phi^{i}_{K})\nabla_{\theta}(\phi^{i}_{K})  \nonumber \\
                          &  = \frac{1}{M}\sum_{i=1}^{M} \nabla_{\phi_{i}} g_{i}(\phi^{i}_{K})\nabla_{\phi^{i}_{K-1}} (\phi^{i}_{K})\cdots \nabla_{\phi^{i}_{0}}(\phi^{i}_{1})\nabla_{\theta}(\phi^{i}_{0}) \nonumber \\
                          &  = \frac{1}{M}\sum_{i=1}^{M} \nabla_{\phi_{i}} g_{i}(\phi^{i}_{K})\prod_{k=1}^{K}\nabla_{\phi^{i}_{k-1}}(\phi^{i}_{k})\cdot\mathbf{I}
\end{align}
In the last step $\nabla_{\theta}(\phi^{i}_{0})\!=\!\mathbf{I}$ which is due to $\phi^{i}_{0}\!=\!\theta$  since the gradient descent starts from the shared parameter $\theta$.

We then expand each $\phi^{i}_{k}$ with its gradient descent update in Eq. \eqref{eqn:maml_sgd}
\begin{align}
\nabla_{\theta} f(\theta) &  \approx \frac{1}{M}\sum_{i=1}^{M} \nabla_{\phi_{i}} g_{i}(\phi^{i}_{K})\prod_{k=1}^{K}\nabla_{\phi^{i}_{k-1}}[\phi^{i}_{k-1} - \alpha \nabla_{\phi_{i}} g_{i}(\phi^{i}_{k-1})] \nonumber \\
                          & = \frac{1}{M}\sum_{i=1}^{M} \nabla_{\phi_{i}} g_{i}(\phi^{i}_{K})\prod_{k=1}^{K}[\mathbf{I} - \alpha\nabla^{2}_{\phi_{i}} g_{i}(\phi^{i}_{k-1})] \nonumber \\
                          & \approx \frac{1}{M}\sum_{i=1}^{M} \nabla_{\phi_{i}} g_{i}(\phi^{i}_{K})
\end{align}
where in the last step we apply the first-order approximation by assuming the second-order Hessian matrix is zero
\begin{align}
    \nabla^{2}_{\phi_{i}} g_{i}(\phi^{i}_{k-1}) = \bm{0} \label{eqn:zero_hessian}, \ \ \ k  \in [K].
\end{align}
Particularly, if we only conduct gradient descent on the lower level for just one step (i.e., $K=1$), we have
\begin{align}
     \nabla_{\theta} f(\theta) \approx \frac{1}{M}\sum_{i=1}^{M} \nabla_{\phi_{i}} g_{i}(\phi^{i}_{1}).
\end{align}

\section{Implementation}
\label{sec:impl}

\setlength{\textfloatsep}{0pt}
\begin{algorithm}
    \caption{Self-Supervised Pre-Training with Local Constraints (PTLOC)}
    \label{alg:ptec}
     \textbf{Input:} data sources $M$, \  iterations $T$, \ local update steps $K$,  \ local learning rate $\alpha$, \ global learning rate $\beta$.

     \vspace{0.2cm}
     Initialize model parameters $\theta_{0}$;

     \tcp{$T$ iterations}
     \For {$t = 1 : T$}
     {
         \tcp{lower level optimization}
         \For {$i = 1 : M$}
         {

            Copy the model from upper level $\phi^{i}_{0} = \theta_{t-1}$; \\

            Sample a batch from $\mathcal{D}_{i}$; \\

            \For { $k = 1 : K$ }
            {
                Compute gradient $\nabla_{\phi_{i}} g_{i}(\phi^{i}_{k-1})$ on this batch;

                Update local model $\phi^{i}_{k} \leftarrow \phi^{i}_{k-1} - \alpha \nabla_{\phi_{i}} g_{i}(\phi^{i}_{k-1})$;
            }
        }
        \tcp{upper level optimization}
        Compute global gradient $\nabla_{\theta} f(\theta_{t-1})=\frac{1}{M}\sum_{i=1}^{M} \nabla_{\phi_{i}} g_{i}(\phi^{i}_{K})$; \\

        Update global model \\
        \hspace{4pt} $\theta_{t} \leftarrow \theta_{t-1} - \beta \nabla_{\theta} f(\theta_{t-1})$;
     }
\end{algorithm}

Based on the derivation in Secs. \ref{sec:lowerp} and \ref{sec:upperp}, the pseudo-code implementation of PTLOC is given in Algorithm \ref{alg:ptec}. The PTLOC training is carried out in $T$ iterations, where the problems of the lower and upper levels are alternately optimized. Specifically, in each iteration, the $M$ lower-level problems on each data source are first (approximately) solved individually using $K$-step gradient descent. The gradient descent is initialized with the global model parameter from the upper level, which is the same for each of the $M$ problems. After the lower-level problem is (approximately) solved, the resulting local optimum of each problem is used to evaluate its gradient. The upper-level problem is then optimized using gradient descent based on the averaged gradients evaluated from the $M$ lower-level problems. We find that in order for PTLOC to perform well, it needs to be appropriately initialized. In our experiments, we always initialize PTLOC with a model trained from CSSL. Based on the observation in \cite{Sagun_EigenHessian}, when a deep model is sufficiently trained, the majority of the eigenvalues of the Hessian matrix of the loss function tend to be zero. This makes the assumption in Eq. \eqref{eqn:zero_hessian} more accurate and therefore the first-order approximation more legitimate.

When considering the first-order approximation with one step gradient descent, PTLOC shares similarity with MAML from the optimization perspective, as MAML can also be interpreted from a bilevel optimization framework \cite{Fan_SignMAML,Abbas_SharpMAML}.  There are works using MAML in acoustic modeling \cite{Hsu_MAMLASR, Zhou_MetaTransfer, Anoop_MAMLIndian,Lin_MAMLSpeaker}, most of which are on supervised learning. However, there are differences between the two: i) PTLOC is on self-supervised learning without relying on ground-truth labels; and, ii) PTLOC uses the same training data in both the upper and lower level problems without any validation data as those in the meta learning design \cite{finn2017model,Abbas_SharpMAML}. The similarity between the two is the outcome of the first-order approximation to the solution of a bilevel optimization problem.

\vspace{-0.2cm}
\section{Experiments}
\label{sec:exp}

We evaluate the proposed PTLOC on two sets of experiments. One is a self-supervised English acoustic model pre-training using speech data from multiple domains based on BEST-RQ \cite{Chiu_Bestrq}. The other is a self-supervised multilingual acoustic model pre-training based on contrastive predictive coding (CPC) \cite{oord2018representation}.

\vspace{-0.2cm}
\subsection{Multi-Domain Pre-Training}
\label{sec:exp_domain}

In this experiment, we use English data collected from a broad variety of domains. All the speech signals have a sampling rate of 16KHz. Table \ref{tab:data} gives the details of the domain distribution and the amount of data from each domain. Data is collected from a total of eight domains, including Broadcast News, ViaVoice dictation, AMI meetings, British (GB) English, Librispeech, Australian (AU) English, Hospitality, and Accented English (Asian and Latin). Among these eight domains, data from the first five domains is used for pre-training, while that from the last five is used for downstream FT and test. Therefore, in the downstream FT tasks, two domains are seen in the pre-training, and three domains are unseen. The data is unbalanced in amount. We use this setting to simulate real-world application scenarios. The hours of speech on test sets are shown in the table. Note that Librispeech (clean/other) and Accented (Asian/Latin) both have two test sets.

\setlength{\textfloatsep}{0pt}
\begin{table}[ht]
\begin{center}
\begin{tabular}{ l  c c c} \hline
   Domain                 &    Pre-training      &    Fine-tuning   &   Test       \\ \hline
Broadcast News            &        420h          &        -          &    -          \\
ViaVoice                  &        450h          &        -          &    -          \\
AMI meetings              &         80h          &        -          &    -          \\
GB English                &        183h          &         50h      &   6.2h       \\
Librispeech               &        860h          &        100h      &  5.4h/5.3h  \\
AU English                &         -             &        250h      &   1.3h       \\
Hospitality               &         -             &         40h      &   1.2h       \\
Accented                  &         -             &         40h      &  2.1h/2.4h  \\ \hline
\end{tabular}
\end{center}
\vspace{-0.5cm}
\caption{Speech data from multiple domains used in self-supervised pre-training and downstream fine-tuning/test tasks.}\label{tab:data}
\end{table}

The pre-trained acoustic model is a Conformer model \cite{Gulati_conformer}. The input is 40-dimensional log-Mel spectrogram features and their first and second-order derivatives. Features of every two adjacent frames are concatenated which gives 240-dimensional input vectors. The model has 10 conformer blocks. Each block has 512 hidden units and 8 attention heads of 64 dimensions. The convolution kernel size is 31. The total number of parameters is 70M. The self-supervised training is carried out using BEST-RQ. The masking probability in BEST-RQ is 0.02. The mask span is 20 frames. The masked frames are replaced with Gaussian noise with 0 mean and 0.1 variance. The size of the random codebook is 256.

For the CSSL baseline, we start the training with a learning rate $2e\!-\!4$ for 60 epochs which is then annealed by $\frac{1}{\sqrt{2}}$ every epoch afterward. The training ends after 80 epochs. For PTLOC, we start the training with a local learning rate $\alpha=1e\!-\!4$ and a global learning rate $\beta=1e\!-\!5$ for 40 epochs. They are then annealed by $\frac{1}{\sqrt{2}}$ every epoch afterward synchronously. The training ends after 60 epochs. The lower-level optimization on each data source is conducted in parallel. To deal with the unbalanced data size across different data sources, we make the batch size roughly proportional to the total amount of data to make each data source produce about the same number of batches. In addition, we also make random skipping of batches during the training when a data source has more batches than the others. All training uses the AdamW optimizer.

After pre-training, a linear layer is added to the pre-trained conformer model for FT with labeled data on each of the downstream ASR tasks using the Connectionist Temporal Classification (CTC) \cite{Graves_CTC} criterion. The softmax layer contains 43 output units, including 42 characters and the null symbol. The FT starts with a learning rate $\beta=1e\!-\!4$ for 5 epochs which is then annealed by $\frac{1}{\sqrt{2}}$ every epoch afterward. The FT ends after 15 epochs.

\setlength{\textfloatsep}{0pt}
\begin{table}[tbh]
\begin{center}
\begin{tabular}{ p{1.8cm} p{0.5cm} c p{0.3cm} p{0.3cm} c} \hline
    Model                 &       Hosp        &      Acct      &      AU     &     GB      &    Libri        \\
                          &                   &  asian/latin   &             &             &  clean/other    \\ \hline
\hspace{-0.2cm}CSSL                   &       22.0        &  16.6/17.8     &    35.2     &    27.9     &   12.4/24.5     \\
\hspace{-0.2cm}PTLOC ($K$=1)           &       18.5        &  15.8/17.0     &    24.6     &    24.0     &   11.5/22.7     \\
\hspace{-0.2cm}PTLOC ($K$=3)           &       18.7        &  15.7/17.1     &    26.9     &    24.5     &   11.5/22.7     \\ \hline
\end{tabular}
\end{center}
\vspace{-0.5cm}
\caption{WERs of CSSL and PTLOC on five downstream ASR tasks with different $K$.}\label{tab:asr1}
\end{table}

Table \ref{tab:asr1} compares word error rates (WERs) of CSSL as the baseline and the proposed PTLOC on five downstream ASR tasks. Each task represents an ASR application in a particular domain. We also compare the performance of PTLOC with different $K$-step local updates in the lower-level optimization. It can be observed that PTLOC shows improvements over CSSL on seven test sets across all downstream tasks. It can also be observed that running more local updates ($K$=3) may not necessarily yield better performance, but increases the computational cost. Therefore, we will stick to $K$=1 in the remaining experiments.

In the experiments in Table \ref{tab:asr1}, we use CSSL to initialize PTLOC. This procedure can be iterative where we can use CSSL and PTLOC to perform mutual initialization. By doing this, both landscapes of the upper global loss and lower local loss will be further explored and hopefully we can end up with a better optimum.  The results are given in Table \ref{tab:asr2}. We conduct three rounds of CSSL and PTLOC sequentially (denoted CSSL.$i$ and PTLOC.$i$, respectively, $i=1,2,3$). We initialize PTLOC.$i$ using CSSL.$i$ and initialize CSSL.$i\!+\!1$ using PTLOC.$i$. In every round, we use the same training schedule as that in Table \ref{tab:asr1}. The results clearly show that this iterative training strategy can greatly improve the adaptivity of the pre-trained model which obtains significant WER reduction in all downstream ASR tasks. The results also show that CSSL does not always improve over the PTLOC model which initializes it on all downstream domains. This is because the averaged global loss can not guarantee good performance on each specific domain. On the other hand, PTLOC always outperforms its initial CSSL model in every downstream domain and hence demonstrates its superior robustness.  Overall, if we compare the final WERs after the iterative training (the last row) to the baseline (the first row), that gives rise to \textbf{15\%-40\%} relative improvements across the seven test sets and the improvements are consistent.  Note that the WERs of the Librispeech baseline (12.4/24.5) use only 100 hours of labeled data for supervised FT. It should not be confused with WERs using 960 hours of labeled data, commonly reported in the literature. Furthermore, its distribution is further flattened by data from various domains. (As a reference, if we only use 860 hours of unlabeled speech and remove data from other domains in CSSL.1, the WERs are 7.4/20.1).

\begin{table}[tbh]
\begin{center}
\begin{tabular}{ l  c c c c c} \hline
    Model                 &       Hosp        &      Acct      &      AU     &     GB      &    Libri        \\
                          &                   &  asian/latin   &             &             &  clean/other    \\ \hline
   CSSL.1                 &  \textbf{22.0}    &  \textbf{16.6/17.8}     &    \textbf{35.2}     &    \textbf{27.9}     &   \textbf{12.4/24.5}     \\
   PTLOC.1                 &       18.5        &  15.8/17.0     &    24.6     &    24.0     &   11.5/22.7     \\ \hline
   CSSL.2                 &       19.6        &  15.3/16.1     &    28.4     &    21.2     &   10.3/20.1     \\
   PTLOC.2                 &       16.2        &  14.2/15.4     &    24.3     &    20.1     &   10.2/19.7     \\ \hline
   CSSL.3                 &       17.4        &  15.5/16.1     &    24.6     &    18.9     &   10.2/19.8     \\
   PTLOC.3                 &  \textbf{15.6}    &  \textbf{13.7/15.1}     &    \textbf{21.3}     &    \textbf{18.4}     &    \textbf{9.4/18.2}     \\ \hline
\end{tabular}
\end{center}
\vspace{-0.5cm}
\caption{WERs of iterative CSSL and PTLOC on five downstream fine-tuning ASR tasks. In the training, CSSL and PTEC alternately initialize each other.}\label{tab:asr2}
\end{table}

\vspace{-0.5cm}

\subsection{Multilingual Pre-Training}
\label{sec:exp_multilingual}

For the multilingual pre-training experiment, we use a subset of the multilingual data from the CoVoST v2 dataset \cite{wang2020covost}, which is sampled at 48 kHz. Table \ref{tab:data_multilingual} provides details on the language distribution and the amount of data for each language. The data includes seven languages. The first three languages (English, French, Dutch) are used for pre-training and the remaining four (Turkish, Swedish, Tamil, Welsh) are used for downstream FT ASR tasks. Unlike the multi-domain experiment in Sec.\ref{sec:exp_domain}, there is no overlap between the pre-training and FT languages in this experiment. But both the pre-training and FT data are unbalanced in quantity.

\setlength{\textfloatsep}{3pt}
\begin{table}[tbh]
\begin{center}
\begin{tabular}{ l  c c c} \hline
    Language                 &    Pre-training      &    Fine-tuning   &   Test       \\ \hline
English            &        357h          &     -             &      -        \\
French                  &        180h          &     -             &       -       \\
Dutch             &         119h          &      -            &      -        \\

Turkish               &          -       &        2h      &  2h  \\
Swedish                &           -           &        2h      &   2h       \\
Tamil               &             -         &         2h      &   1h       \\
Welsh                  &           -           &         1h      &  1h  \\ \hline
\end{tabular}
\end{center}
\vspace{-0.5cm}
\caption{Multilingual speech data used in self-supervised pre-training and downstream fine-tuning/test tasks.}\label{tab:data_multilingual}
\end{table}

We pre-train a Conformer model using raw audio. We employ a 1D convolutional layer with a kernel size of 3 to capture local temporal dependencies in the input signal before passing it to the Conformer encoder. During pre-training, the CPC loss is computed using a context length of 256 samples, along with 12 positive and 12 negative samples per instance. The model consists of 8 Conformer blocks, each containing 512 hidden units and 8 attention heads of 64 dimensions. The convolutional kernel size is 31. The total number of parameters is 59M.

For the CPC baseline, we start the training with a learning rate of $5e\!-\!3$ and continue training for 80 epochs which is then annealed by $\frac{1}{10}$ every epoch for next 20 epochs. The training ends at 100 epochs. For PTLOC, we start the training with a local learning rate $\alpha\!=\!5e\!-\!3$ and global learning rate $\beta\!=\!5e\!-\!4$ for 60 epochs. They are then annealed by $\frac{1}{10}$ every epoch afterward synchronously. The training ends after 80 epochs. All the training uses the AdamW optimizer.

After pre-training, a linear classification layer is added to the pre-trained Conformer model for FT on labeled data for each downstream ASR task using the CTC criterion. The softmax layer comprises 1,000 labels, which are generated using SentencePiece \cite{kudo2018sentencepiece}. FT is performed for 30 epochs with an initial learning rate of \( \beta\!=\!\!5e\!-\!5\), which is reduced by a factor of 10 after each subsequent epoch. The FT concludes after 50 epochs.

\setlength{\textfloatsep}{0pt}
\begin{table}[tbh]
\begin{center}
\begin{tabular}{ l  c c c c} \hline
    Model                 &       Turkish        &      Swedish      &      Tamil     &     Welsh               \\ \hline
   CSSL.1                 &  \textbf{41.6}    &  \textbf{52.1}     &    \textbf{72.2}     &    \textbf{57.9}         \\
   PTLOC.1                 &       40.2       &  51.3    &    71.6    & 57.1    \\ \hline
   CSSL.2                 &       39.9        &  50.8     &    71.5       &   56.8     \\
   PTLOC.2                 &       38.2        &  48.9     &    70.7     &     56.1     \\ \hline
   CSSL.3                 &       38.1        &  48.5     &    70.3     &     56.0     \\
   PTLOC.3                 &  \textbf{37.6}    &  \textbf{47.8}     &    \textbf{70.1}     &    \textbf{55.2}    \\ \hline
\end{tabular}
\end{center}
\vspace{-0.5cm}
\caption{WERs of iterative CSSL and PTLOC on four downstream fine-tuning ASR tasks on different languages. In the training, CSSL and PTLOC alternately initialize each other.}\label{tab:asr_multilingual}
\end{table}

Table~\ref{tab:asr_multilingual} compares the WERs of CSSL, used as the baseline, and the proposed PTLOC across four downstream ASR tasks in different languages. We also follow the same iterative mutual initialization process as in the multi-domain experiments. We conduct three rounds of sequential training, where CSSL is trained for 100 epochs, followed by 80 epochs of PTLOC training in each round. The results from the multilingual experiments exhibit a similar trend to those observed in the multi-domain setting. The proposed iterative training strategy significantly enhances the adaptivity of the pre-trained model, leading to consistent WER reductions across all four unseen languages. Specifically, when comparing the final WERs after iterative training (the last row) to the baseline (the first row), we observe relative improvements of \textbf{2.9\%–9.6\%} across the four test sets.

\section{Summary}
\label{sec:sum}

In this paper, we propose PTLOC to deal with data heterogeneity in self-supervised pre-training. Local constraints are imposed to ensure that the models optimize each data source to its local optimum after K-step gradient descent initialized from the model. We use the first-order approximation to solve the resulting bilevel optimization problem. Experiments are carried out on multi-domain and multilingual ASR tasks. It shows that PTLOC can significantly improve the adaptivity of the pre-trained model, which can yield improved performance in downstream fine-tuning tasks.

\bibliographystyle{IEEEbib}
\bibliography{ptloc}

\end{document}